\documentclass[sigconf]{acmart}

\usepackage{balance}

\AtBeginDocument{%
  }

\author{Taras Kucherenko}
\orcid{1 0000-0001-9838-8848}
\affiliation{%
  \institution{SEED, Electronic Arts}
  \city{Stockholm}
  \country{Sweden}
}
\email{tkucherenko@ea.com}

\author{Derek Peristy}
\affiliation{%
  \institution{Create, Electronic Arts}
  \city{Vancouver}
  \country{Canada}
}
\email{derekp@ea.com}

\author{Judith B{\"u}tepage}
\affiliation{%
  \institution{SEED, Electronic Arts}
  \city{Stockholm}
  \country{Sweden}}
\email{jbutepage@ea.com}

\renewcommand{\shortauthors}{Taras Kucherenko, Derek Peristy, and Judith Bütepage}

\copyrightyear{2025}
\acmYear{2025}
\setcopyright{acmlicensed}\acmConference[MM '25]{Proceedings of the 33rd ACM International Conference on Multimedia}{October 27--31, 2025}{Dublin, Ireland}
\acmBooktitle{Proceedings of the 33rd ACM International Conference on Multimedia (MM '25), October 27--31, 2025, Dublin, Ireland}
\acmDOI{10.1145/3746027.3755788}
\acmISBN{979-8-4007-2035-2/2025/10}

\begin{document}

\title[Evaluating the Evaluators]%
      {Evaluating the Evaluators: Towards Human-aligned Metrics for Missing Markers Reconstruction}



\begin{abstract}
  Animation data is often obtained through optical motion capture systems, which utilize a multitude of cameras to establish the position of optical markers. However, system errors or occlusions can result in missing markers, the manual cleaning of which can be time-consuming. This has sparked interest in machine learning-based solutions for missing marker reconstruction in the academic community. Most academic papers utilize a simplistic mean square error as the main metric. In this paper, we show that this metric does not correlate with subjective perception of the fill quality. Additionally, we introduce and evaluate a set of better-correlated metrics that can drive progress in the field. 

\end{abstract}



\begin{teaserfigure}
 \includegraphics[scale=0.53]{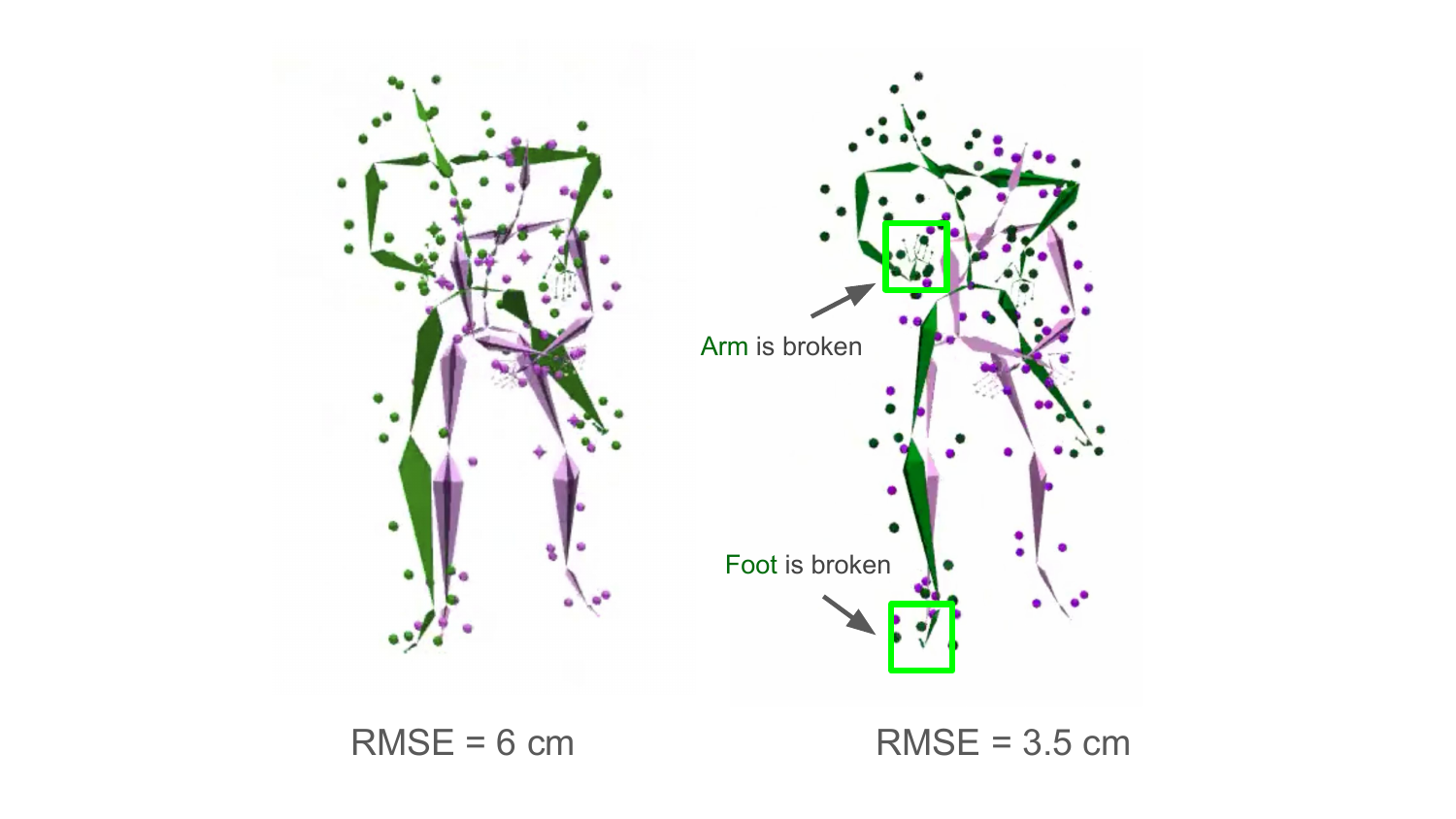}
 \centering
  \vspace{-5mm}
  \caption{Motion capture data from two actors after filling gaps and solving into skeletons. The left and right sides show two different gap-filling methods. This illustration highlights a major issue with commonly used mean squared error metrics for reconstructing missing markers. Although the left prediction looks visually accurate, it shows a higher error than the clearly flawed pose on the right, with broken parts highlighted in light green. See the video attachment for an animation example.}
  \vspace{1mm}
\label{fig:teaser}
\end{teaserfigure}


\maketitle

\section{Introduction}
\label{sec:intro}
Optical motion capture (MoCap) is a process that utilizes a multitude of cameras to establish the position of optical markers. This process is followed by fitting a skeleton into the point cloud of optical markers' positions. The latter operation is commonly referred to as \emph{solving}. MoCap technology is extensively utilized in the film and video game industry to animate virtual characters with lifelike movements. Although marker-based solutions capture body movements with more precision than purely video-based trackers, even state-of-the-art MoCap equipment and software suffer from occasional marker occlusion and mislabeling, as well as positional errors. Manual clean-up is therefore still necessary. As this process can be particularly time-consuming, even for experienced animators, the idea of automating MoCap data cleaning has long been of interest~\cite{liu2006estimation}. Machine learning (ML) solutions, and in particular deep learning methods, have been proposed to solve this problem, e.g.~\cite{holden2018robust, perepichka2019robust, kaufmann2020convolutional, cui2021towards, xia2022local, lee2024denoising}.

Although new ML methods are proposed regularly, little focus has been given to evaluating and benchmarking MoCap cleaning systems. Current evaluation practices present two main problems. 

First, papers vary in terms of datasets and evaluation protocol. The general setup of related papers is to define an artificial distribution over missing markers and positional errors and to corrupt both the training and the testing data using this distribution~\cite{holden2018robust}.  Approaches vary in terms of distributions of missing markers during test time. While some assume individual markers to be missing independently from each other \cite{perepichka2019robust, chen2021MoCap, yang2024u}, others assume gaps over different time windows \cite{kucherenko2018neural, li2021graph, skurowski2024tree} or body parts \cite{cui2021towards, lohit2021recovering, perepichka2019robust}. The performance measured in these papers thus only indicates how well the model would perform on data corruptions that are distributed similar to the artificial distribution, not how well it would perform on real-world MoCap recordings~\cite{pan2023locality}.
We evaluate models using real-world data to assess their performance in a realistic setting.

The second problem of current evaluation practices is that metrics focus on reconstruction errors and not on aspects that are important to the end user of the data, in this case animators. The predominant metric in related work is the mean squared error (MSE) between the systems predictions for missing markers and the ground truth position \cite{holden2018robust, perepichka2019robust, xia2022local, cui2019deep, lohit2021recovering, li2021graph, kieu2022locally, cai2021unified, yang2024u}. However, MSE-like metrics fail to measure spatial and temporal correlation. Both temporal consistency and consistent bodily proportions are important to animators because they make the solving to a skeleton and manual editing easier. For this reason, spatial and temporal consistency are more important than the distance between the predicted sequence and a ground truth sequence. We, therefore, propose metrics that align with animators' needs and assess how well different metrics correlate with domain expert judgment of reconstruction quality. 

To verify that metrics should take spatial and temporal consistency into account, we propose and evaluate such metrics, namely Bone Distance Preservation (BDP) and Velocity Distance (VD). To see if these metrics align with the perceptual quality, we correlate metric values with human judgment of animation quality. To this end, we perform a user study in which domain experts are asked to rate motion capture sequences after they have been cleaned by different ML systems and solved onto a skeleton. We then analyse correlations between human judgment and different metrics.

It is important to note that this work focuses on the evaluation of ML-based MoCap cleaning systems and not on presenting a novel, state-of-the-art system. Therefore, we do not benchmark our models on publicly available datasets. 

Furthermore, while we do propose new metrics and modifications of existing metrics, the main point of this paper is not to provide a solution but rather to outline a problem with the commonly used metrics in the field.

In conclusion, our contributions are:
\begin{itemize}
   \item Conducting the first human-subject study to assess the perceived quality of predicted markers.
   \item Developing new metrics to evaluate missing marker reconstruction solutions, including those that do not require access to ground truth data.
   \item Performing a correlation analysis between subjective and objective ratings to validate various metrics.
\end{itemize}

\section{Related work}
\label{sec:related}

This paper primarily focuses on evaluation practices. Therefore we only briefly touch upon different modeling approaches to solve the missing marker problem (Section \ref{sec:related:modeling}) and dedicate more attention to metrics (Section \ref{sec:related:metrics}). For a detailed review of the missing markers reconstruction field we refer the reader to \cite{martini2023denoising}. 

\begin{figure*}[ht]
\centering
  \includegraphics[width=0.9\linewidth]{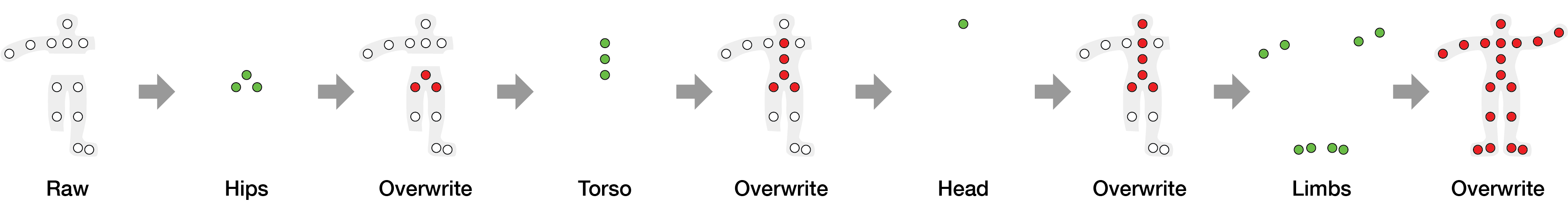}
  \caption{Hips outwards model illustration. Markers are filled one body part at a time in the following order: hips -> torso -> head -> limbs.}
  \label{fig:hip_outwards_model}
\end{figure*}

\subsection{Modeling}
\label{sec:related:modeling}

There are two main ideas that dominate the deep learning based approaches to solving the missing marker problem. The first is to design a model with an encoder - decoder structure \cite{holden2018robust, kucherenko2018neural, cui2019deep, kaufmann2020convolutional, lohit2021recovering, xia2022local}. This is motivated by the observation that the modeling of human motion only requires a small number of dimensions \cite{wang2005gaussian}. The mapping to a lower dimensional space reduces the problem of filling in arbitrarily missing markers to a nearest neighbor search in said space. The second theme that unites most deep learning approaches is to view the missing marker task as a sequence modeling problem. In deep learning, sequences are commonly modeled using convolutions networks \cite{kaufmann2020convolutional, cui2021towards, lohit2021recovering, xia2022local}, recurrent neural networks such as long short-term memory models  \cite{kucherenko2018neural, cui2019deep, li2021graph} and, more recently, attention-based solutions \cite{cui2019deep, cui2021towards, cai2021unified, yang2024u}. 


Finally, we can differentiate between approaches that focus solely on missing marker reconstruction, e.g. \cite{kucherenko2018neural, cui2019deep, lohit2021recovering}, and approaches that solve from raw data to a character, e.g. \cite{ holden2018robust, perepichka2019robust, pan2023locality, yang2024u}. In the latter case, the usual approach is to recover the marker positions before solving to a skeleton.

\subsection{Metrics}
\label{sec:related:metrics}

We here focus on metrics computed on a marker level. Previous work that combined MoCap cleaning and solving to a character sometimes use similar metrics but on the character level instead of the marker level \cite{ holden2018robust, perepichka2019robust, pan2023locality, yang2024u}.

Most of the previous work in the field used some form of the mean square error (MSE). Different names are used such as position error \cite{holden2018robust, perepichka2019robust, xia2022local}, Root Mean Square Error (RMSE) \cite{cui2019deep, lohit2021recovering, li2021graph}, MSE \cite{kieu2022locally}, Mean Per Joint Position Error (MPJPE) \cite{cai2021unified, yang2024u}, and more. Fundamentally they do not differ significantly because they all measure the distance between the predicted and the ground truth sequences.

MSE-based metrics are problematic because they do not account for spatial and temporal cohesion. For example, if a marker is 1 cm off in one direction in this frame and 1 cm off in the opposite direction in the next frame - the data is  broken despite a small MSE loss. Similarly, if one hip marker is 1 cm up and the other hip marker is 1 cm down - the hip structure is broken,  but the MSE loss is small. On top of that, it has been shown that MSE does not correlate with subjective perception of motion quality is other animation fields, such as text-to-motion prediction \cite{voas2023best} or gesture generation \cite{kucherenko2023evaluating}.

One possible way to take spatial cohesion into account is to calculate bone distance preservation distance \cite{cui2019deep, kaufmann2020convolutional, yin2021graph}. We experiment with different ways to calculate bone distance preservation in this paper.

One way to take temporal cohesion into account is to calculate velocity distance. This has not been done before in the field of missing markers reconstruction, to the best of our knowledge, and we propose two novel metrics based on velocity distance which are detailed in Section \ref{ssec:objective_metrics}.



\section{Methodology}
\label{sec:method}

As mentioned in the introduction, the focus of this paper is not to propose a new state-of-the-art solution. Our main objective is the evaluation of such models (see Section~\ref{sec:exp_setup}). Keeping this in mind, here we present the data and modeling choices we have made to solve the missing marker problem.

\subsection{Problem formulation}
\label{subsec:problem_formulation}

Let $Y \in \mathcal{R}^{(T, 3 N_m)}$ be a matrix that represents a sequence of $T$ frames and a skeleton that consists of $N_m$ markers. We distinguish between raw observations $Y_{raw}$ and cleaned observations $Y_{clean}$. Further, we use dot $\dot{Y}$ to denote a sequence that has been transformed by noise and hat  $\hat{Y}$ to denote a predicted sequence.

The usual problem statement of missing markers reconstruction problem is to assume a ground truth marker sequence $Y_{clean}$. In order to simulate marker occlusion and positional noise, $Y$ is transformed with the help of a masking matrix $O  \in [0,1]^{(T, 3 N_m)}$ and a noise matrix $\Sigma  \in \mathcal{R}^{(T, 3 N_m)}$ as follows
\begin{equation}
    \dot{Y}_{clean}  = (Y_{clean}  + \Sigma) * O.
\end{equation}
The masking matrix usually consists of independent and identically distributed (iid) Bernoulli variables with a certain probability of a given matrix entry being zero (occluded) or one (observed). The noise matrix $\Sigma$ follows an iid Gaussian distribution with zero mean and a small variance. Sometimes it might be multiplied with its own masking matrix to control how many markers are shifted at every training step. This distribution is applied both during training and testing due to the lack of real-world raw data. This problem setting can be seen as a mixture of denoising and reconstruction.    

We approach the problem slightly differently by converting it into a pure denoising task. Filling missing values with zeros is always tricky as this changes the distribution of the data drastically. It can confuse the model as zeros might be very close to actually observed small values. For that reason, we convert the reconstruction problem into a pure denoising problem. During training, we add iid Gaussian noise to clean data but only for those entries that have been sampled to be missing:
\begin{equation}
    \dot{Y}_{clean}  = Y_{clean} + \Sigma * (1-O).     \label{eq:ournoise_train}  
\end{equation}
 
We then train the model to denoise the marker values that were randomly shifted. During testing, instead of filling the missing marker sequences with zeros, we  
use cubic interpolation between the last observation of a marker and the first observation following the gap:

\begin{align}
    \dot{Y}_{raw}  &= \text{interpolate}(Y_{raw}).   \label{eq:ournoise_test} 
\end{align}

The task of the ML model is to predict the correct position of the missing markers:
\begin{equation}
    \hat{Y}  = f(\dot{Y}).  \label{eq:pred}  
\end{equation}

Since the model has been trained to shift markers to the correct position, it is able to shift the interpolated values correctly.

\subsection{Models considered}

This paper focuses on evaluation rather than modeling. We do not present a new model and do not claim to be the new state-of-the-art. Instead, we evaluate two simple models based on commonly used one-dimensional Convolutional Neural Network (CNN) \cite{kaufmann2020convolutional, lohit2021recovering, xia2022local}.

The first model is straightforward, consisting of a sequence of transformations: one 1D CNN layer with LeakyReLU activations and dropout and a final linear layer. We refer to this model as \textbf{Vanilla CNN}. 

The second model takes a structural approach instead. In contrast to previous structural modeling approaches that make use of graph neural networks \cite{cui2021towards, yin2021graph, lee2024denoising}, we introduce structure mainly during inference by replacing missing markers one body part at a time. Our model consists of four modules, each with the same architecture as the first model, but each having different hyper-parameters and different network weights. Instead of predicting all markers at once, each module predicts one body part at a time. As shown in Figure \ref{fig:hip_outwards_model}, it follows the order often used for manual cleanup: hips -> torso -> head -> limbs. 
Prediction is done iteratively. 
In each step, we first predict values for the current body part and then overwrite the interpolated or noisy values in $\dot{Y}$ with the corresponding predicted values: 
\begin{align}
\hat{Y^{i}} &= f^i(\dot{Y}^{i-1}), \\
\dot{Y}^{i} &= \dot{Y}^{i-1} * (1 - B^i) + \hat{Y}^{i} * B^i.  
\end{align}
where iteration $i = [1:4]$ indicates the body part in [hips, torso, head, limbs], $f^i$ is the corresponding body part predictor module, and $B^{i}$ is a masking matrix that consists of zeros except for the row entries that correspond to the missing markers forming the body part $i$. $\hat{Y^{i}}$ is the model prediction for the body part  $i$. $\dot{Y}^{0}$ simply equals the initial $\dot{Y}$ and the final prediction is given by $\hat{Y} = \dot{Y}^4$. 
We call this model \textbf{Hip Outwards}.

\subsection{Training setup}
\label{sec:method:setup}

We train our model as a denoising autoencoder, similar to some prior work \cite{holden2018robust,kucherenko2018neural}. The clean motion capture sequence is first corrupted by adding masked Gaussian noise $\Sigma \sim \mathcal{N}(0, I\sigma)$ and then passed through the model, which tries to reconstruct the original sequence. The amount of noise is gradually increased during training: $\sigma = (\tanh((ep - 10) / 20) + 1) * c / 2$, where $ep$ is the epoch number and $c$ is a parameter selected during hyper-parameter search. This function was selected so that noise level starts close to zero and is never larger than $c$. 

To train the models, we use a linear combination of a MSE-loss of positions and a MSE-loss of velocities of missing markers:
\begin{equation}
loss = \frac{1}{T}\sum_{t=1}^{T}o_t\lVert y_t - \hat{y}_t \rVert ^2 + \lambda * \frac{1}{T-1}\sum_{t=2}^{T}o_t\lVert v_t - \hat{v}_t \rVert ^2 \label{eq:loss}
\end{equation}
where $T$ is the number of frames in the sequence evaluated, $y_t$ is the $th$ column in the ground truth matrix $Y$ corresponding to time step $t$, $\hat{y}_t$ is the $th$ prediction, $v_t$ is the $th$ ground truth velocity, $\hat{v}_t$ is the velocity of the prediction, and  $o_t$ is the $th$ entry of the missing marker mask. Velocity is calculated using the 1st order numerical derivative: $v_t = y_t - y_{t-1}$. $\lambda$ is a multiplier established experimentally to balance the two loss terms. The velocity loss ensures that the filled markers are smoothly connected to existing markers.

During training, we emulate gaps by gradually increasing the number of markers missing and the duration of the gap. This is in contrast to related work that often uses a fixed percentage of randomly missing markers \cite{holden2018robust,perepichka2019robust, chen2021MoCap, lohit2021recovering, xia2022local, yang2024u}. The increase of missing markers follows $number = ep * N_{rate} + N_{start}$ , where $ep$ is the epoch number and $N_{start}$ and $N_{rate}$ control the initial number of missing markers and the rate at which this number increases. Similarily, the duration increases as $duration = ep * D_{rate} + D_{start}$. $number$ and $duration$ are rounded to the closest integer and are capped at the number of existing markers and the sequence duration, respectively. $N_{start}, N_{rate}, D_{start} \text{ and } D_{rate}$ are selected during hyper-parameters search. The sampled gaps are represented in the missing marker mask $O$ that is used to compute the loss in Equation~\ref{eq:loss}. This ensures that 1) the gradient is only propagated for the missing markers; 2) the filled markers are smoothly connected to existing markers.

\section{Experimental setup}
\label{sec:exp_setup}
This section outlines the experimental setup. We work with a real-world dataset used for animations. Our aim is to measure the correlation between different metrics and human perception of reconstruction quality. Here we describe the objective metrics used in this work as well as the setup of our user study. 

The objective metrics are computed on a marker level while we use animations solved to a character in the user study. We chose to use the two different representations over either computing the metrics on a solved character or asking humans to evaluate marker sequences. On the one hand, evaluating the model on the marker level correctly reflects the performance of the model in the data space that it is operating in. Without appropriate metrics in marker space, one would have to continuously have to solve the predictions during model development, which is not trivial. On the other hand, humans, and even domain experts, struggle to evaluate the quality of point clouds. Asking them to evaluate animations on a solved skeleton will, therefore, result in more accurate estimates of quality.  


\subsection{Dataset}

This study investigates the practical utility of objective metrics. In order to do that, we step away from the commonly used synthetic academic datasets such as CMU MoCap \cite{CMU} and work with real-world data used for animation purposes. 


Our internal dataset contains recordings of two actors emulating various close-combat actions, mainly grappling. Due to occlusions by the other actor, it does, therefore, contain large gaps. On average, gaps consist of 25 \% of markers for on average of 1 second. Each actor has 63 markers, for a total of 126 markers. The data was recorded at 120 frames per second (fps). In total, we had access to 230,365 frames meaning roughly 30 minutes of motion data. Those were split in the following way: 217,318 for training, 6,263 for validation, and 6,784 for testing.

The model predictions were post-processed by overwriting the existing markers with their original values and smoothing the sequence using the Savitzky–Golay filter \cite{savitzky1964smoothing}.

\subsubsection{Data processing}
Prior to training the dataset was pre-processed in the following way.

The data was hips-centered by removing an average of the hips' coordinates from each actor's data at every frame, so that the hips center for each actor is at the origin (0,0,0) at every frame. The distance between the actors was stored separately at each time frame since this information is lost in the hips-centering step. The input to the model for a sequence of length $T$ is thus of size $[T, 2 * 63 * 3 + 3]$.

The dataset was augmented by mirroring in the x-axis, mirroring in the y-axis (the z-axis is the vertical axis), and swapping the two actors' places.

During training only Gaussian noise is added to the data. During testing on raw data, the missing values were replaced by interpolation as described in Section~\ref{subsec:problem_formulation}. All the markers were present in the first and last frame at every sequence (which is usually a T-pose) to make sure that no extrapolation is needed.

\subsection{Objective metrics}
\label{ssec:objective_metrics}

Our aim is to establish a set of metrics that evaluate different aspects of marker reconstruction quality. 
We evaluate our models using the following metrics, which we describe in detail below: 
\begin{enumerate}
    \item Mean squared error of ground truth and predicted markers 
    \item Bone distance preservation of predicted with respect to ground truth markers
    \item Bone distance preservation of predicted markers across time steps
    \item Velocity error of predicted with respect to ground truth markers 
    \item Velocity error of predicted markers across time steps 
\end{enumerate}

\subsubsection{Mean squared error}

Every paper in the field employs some version of the mean square error among the evaluation metrics, often as the only evaluation metric \cite{holden2018robust, perepichka2019robust, xia2022local,cui2019deep, lohit2021recovering, li2021graph,kieu2022locally, cai2021unified, yang2024u}.

We here use the Root Mean Square Error (RMSE) which is defined as the name suggests:
\begin{equation}
\sqrt{ \frac{1}{T}\sum_{t=1}^{T}\lVert y_t - \hat{y}_t \rVert ^2 }
\end{equation}

RMSE does not take into account temporal or spatial cohesion, which are both important for the motion capture data.

\subsubsection{Bone distance preservation}
To evaluate spatial cohesion sometimes bone distance preservation (BDP) is used as an additional metric \cite{cui2019deep, kaufmann2020convolutional, yin2021graph}. There are two ways to define what is meant by preserving bone distance: either we measure whether bone distance is preserved with respect to the ground truth or with respect to the previous time frame. The former requires access to the ground truth, while the latter does not. Hence we will call the two \textit{BDP with GT} and \textit{BDP without GT}

BDP with GT was previously used \cite{cui2019deep, kaufmann2020convolutional, yin2021graph} and is defined as:
\begin{equation}
\sqrt{\frac{1}{T * D} \sum_{t=1}^{T} \sum_{d=1}^{D} (L_{td} - \hat{L}_{td} )  ^2 }
\end{equation}
where $D$ is the number of bones considered, $L \in [TD]$ is the bone length matrix of the ground truth sequence, and $\hat{L} \in [TD]$ is the bone length matrix of the prediction.

BDP without GT was not used before and is defined as:
\begin{equation}
\sqrt{\frac{1}{(T-1) * D}\sum_{t=1}^{T-1} \sum_{d=1}^{D} ( \hat{L}_{{t+1},d} - \hat{L}_{td} )  ^2}
\end{equation}


It is not obvious to define bone length in the 3D point cloud of markers, as none of the markers is placed directly on the bone (that would be very painful). There are two ways to work around this: 1) to solve the MoCap data into a skeleton and then apply the bone length preservation metric; 2) to define heuristics about which markers roughly follow a bone. We follow the second approach as solving marker set into a skeleton is not a part of our approach.

We define bone length as the Euclidean distance between the average positions of the two markers placed on different sides of a bone.
The following five bones are considered \cite{vicon}: head (from left to right), left hand (left elbow to wrist), right hand (right elbow to wrist), hips (left to right), left leg (left knee to ankle), right leg (right knee to ankle). For example, for the left-hand bone distance, we first calculate elbow position coordinates as the midpoint between the external and internal elbow markers. We then calculate wrist position as the midpoint between the external and internal wrist markers. Finally, we calculate the left-hand bone distance as the Euclidean distance between the left elbow position, and left wrist position estimates.

The bone distance preservation metric does take spatial coherence into account, but it does not consider temporal smoothness of the data.

\subsubsection{Velocity distance}
To measure temporal coherence, we propose to calculate a velocity distance (VD). This has been previously used in other animation tasks \cite{karras2017audio, holden2020learned}, but not for missing marker reconstruction. As in the case of bone distance preservation, there are two ways to define a velocity distance: either with respect to the ground truth or with respect to the previous time frame. We will call the two \textit{VD with GT} and \textit{VD without GT}.

The velocity distance with ground truth (VD with GT) is defined as
\begin{equation}
\sqrt{ \frac{1}{(T-1)}\sum_{t=2}^{T} \lVert v_t - \hat{v}_t \rVert ^2 }.
\end{equation}

The velocity distance without ground truth (VD without GT) is defined as
\begin{equation}
\sqrt{ \frac{1}{(T-2)}\sum_{t=2}^{T-1} \lVert \hat{v}_{t+1} - \hat{v}_t \rVert ^2}.
\end{equation}

\subsection{User study setup}
We perform a subjective evaluation through a user study to collect ratings for MoCap sequences, aiming to assess how well the objective metrics are correlated with subjective quality perception.

\begin{figure}[b]
\centering
  \includegraphics[width=0.7\linewidth]{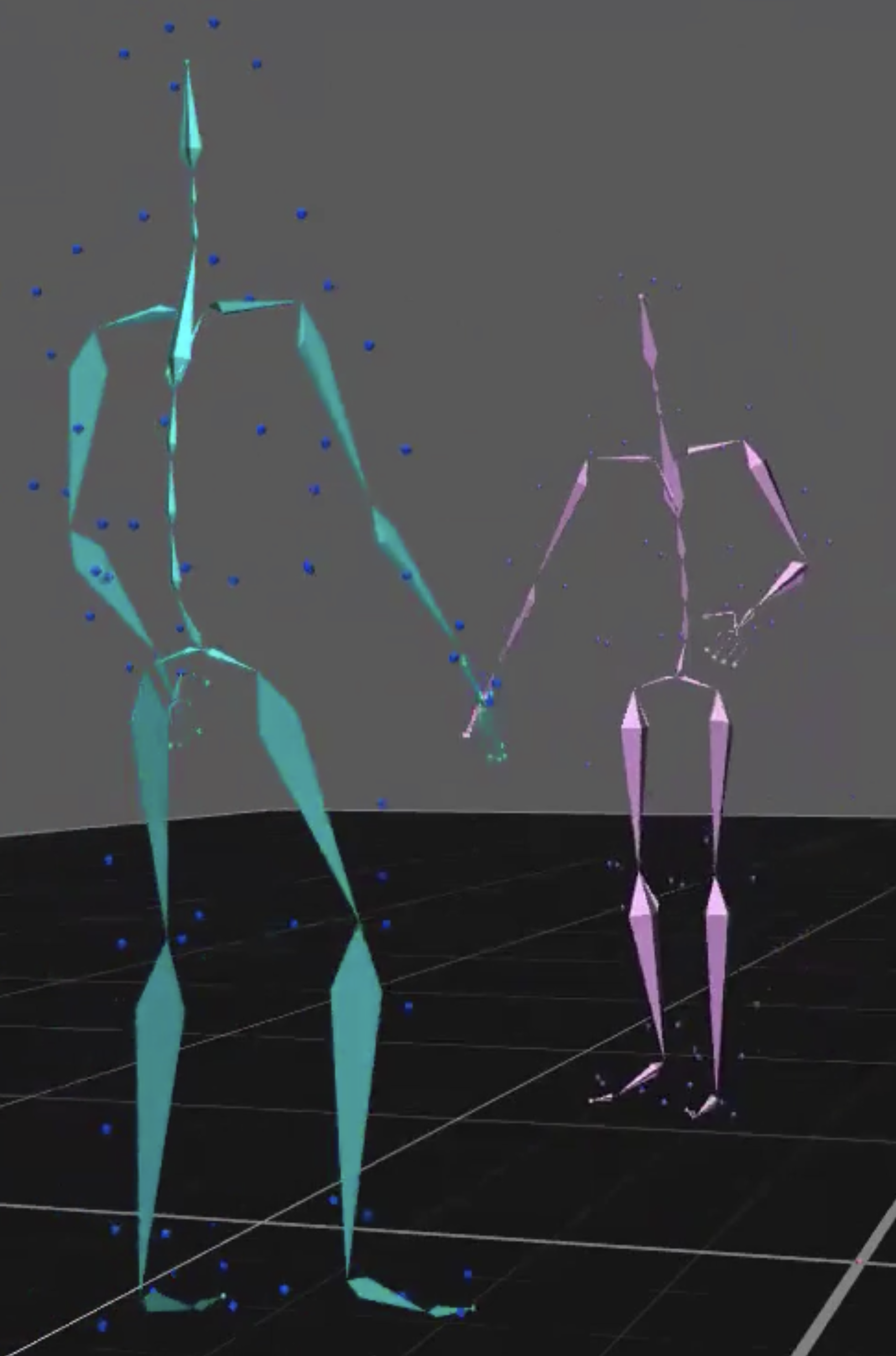}
  \caption{Screenshot from the data provided in the user study.}
  \label{fig:user_study_video_screenshot}
\end{figure}
We use  four conditions in our user study.
The first two conditions consist of the Vanilla CNN and the Hips Outwards model as described above. We also show manually cleaned MoCap data and denote this as the ground truth. The final condition consists of the ground truth to which we add with a small amount of iid Gaussian noise (with 2 cm standard deviation). This noise condition serves as an example of clearly corrupted data.

To obtain user study stimuli, we solve motion capture data to obtain skeletal animation for each segment. That animation is then overlayed on top of the missing markers reconstruction as you can see in Fig. \ref{fig:user_study_video_screenshot}. We motivate the choice to run the user study using solved instead of raw MoCap animations by the fact that very few animators are accustomed to judging MoCap data quality. As the end goal of MoCap cleaning is the usage of the solved MoCap sequence, it makes sense to allow animators to evaluate the quality in this space.

For each condition, we use three different test sequences of approximately 20 seconds. Those sequences are further divided into shorter segments of 4-5 seconds to ensure that each segment contains a single type of movement. Given the primary interest in the first two conditions (the two ML models), we select four segments from each sequence for these. For the ground truth and noisy conditions, we select only one segment from each sequence, namely the middle one with the main part of the motion.
There were 30 stimuli in total: 12 segments from each of the two models considered, three noisy segments, and three clean segments.

Each animator only worked with a subset of the stimuli that was selected randomly to ensure that each stimulus received roughly the same number of responses. The order of stimuli was counterbalanced by shuffling files randomly and inserting one random clean sequence followed by one random noisy sequence at the beginning. This was done to calibrate the range of possible quality levels for the study participants. Each participant would see the two calibration sequences plus one example of each of 12 randomly shuffled sampled from different conditions. In total, each participant was provided with 14 segments to evaluate. 


The stimuli were distributed to professional animators in form of FBX files. This allowed the animators to change view points and replay the sequence as they saw fit. Animators often rate the quality of animations in terms of how much time they would need to bring the animation to a production-ready standard. This inspired us to ask the study participants the following question: 

"How would you rate the usability of this MoCap animation as a starting point?"
\begin{enumerate}
    \item Completely unusable; a new shoot is required.
    \item Usable, but requires extensive manual editing.
    \item Usable with moderate manual editing.
    \item Usable with minimal manual editing
    \item Fully usable as is.
\end{enumerate}

We consider the responses as corresponding numerical values 1-5 and calculate the average of the responses each stimulus received as its subjective perception score.

\begin{figure}[t]
\centering
  \includegraphics[width=0.98\linewidth]{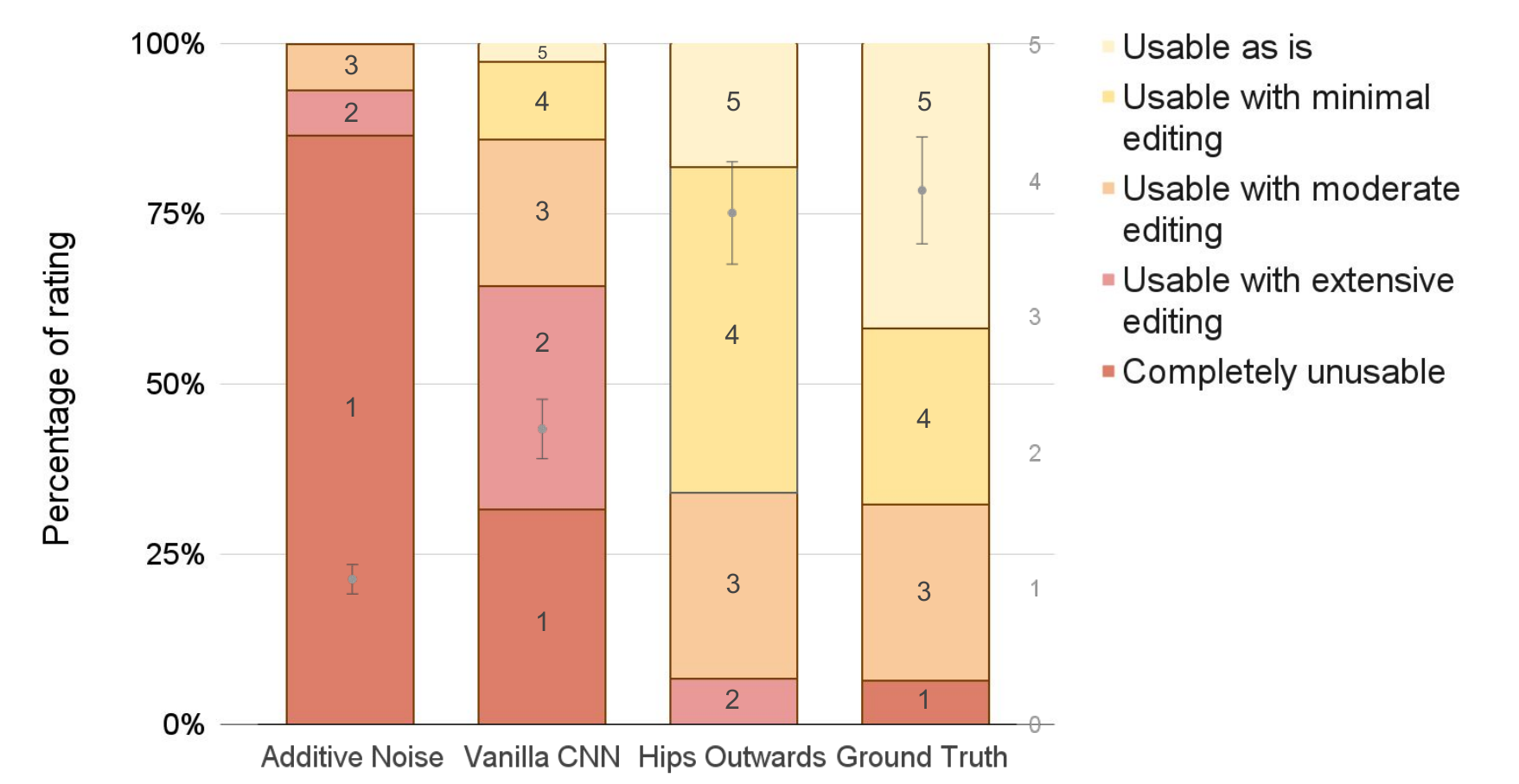}
  \caption{Results of the user study. Fractions of each rating (with y-axis on the left) as well as average scores (with y-axis on the right) with their 95\% confidence intervals.}
  \label{fig:results_user_study}
\end{figure}

\section{Results}

In this section, we present and discuss the results of our subjective and objective evaluation. The main result is our correlation analysis of objective metrics and subjective perception of reconstruction quality. 

\subsection{User study}

\begin{table*}[!th]
    \centering
    \begin{tabular}{|l|c|c|c|c|c|}
    \toprule
  Condition  $\downarrow \quad$  Metric $\rightarrow$ & RMSE & VD with GT & VD & BDP with GT & BDP \\  \midrule
Additive Noise & 1.997 & 4.897 & 4.927 & 1.927 & 2.725   \\  
Vanilla CNN & 13.779 & 0.41 & 0.546 & 4.213 & 0.079    \\    
Hips Outwards & 3.143 & 0.128 & 0.413 & 2.679 & 0.040   \\  
Ground Truth & - & - & 0.519 & - & 0.027  \\ \bottomrule
    \end{tabular}
    \vspace{4mm}
    \caption{The objective metrics listed for each condition in cm. Lower is better. Metrics that compare with the Ground Truth (RMSE, VD with GT and BDP with GT) are shown as missing for the Ground Truth condition as they will always be zero.}
    \label{tab:objective}
    
\end{table*}

Our study involved 15 participants: 11 males and 4 females. The average age was 40.2, with a standard deviation of 7.2. All were professional animators.

The user study results are shown in Figure \ref{fig:results_user_study}. The bars are depicting the percentage value of each rating that each condition (Additive Noise, Vanilla CNN, Hips Outwards and Ground Truth) received. The ratings are color coded from 1 (completely unusable) to 5 (usable as is). We also depict the average rating with 95\% confidence intervals in black.

As expected, we can see that samples in the Additive Noise condition received the lowest possible score, and the Ground Truth sequences received high but not always the highest scores. The two models considered are rated higher than the Additive Noise baseline and lower than the Ground Truth condition. However, on average the Hip Outwards model was rated much higher than Vanilla CNN and nearly as highly as the Ground Truth condition.  This indicates that a structural approach to missing marker reconstruction is helpful.

We also allowed users to add comments. Certain issues were mentioned several times. The most common keywords used in the comment section were 

\begin{enumerate}
    \item foot planting / drifting / floating / sliding;
    \item flipping / flicking / swapping / popping;
    \item jitter / noise / shaking; 
    \item broken / twisted.
\end{enumerate} 

These observations could be used to develop future objective metrics, for example, a foot sliding metric as in \cite{karunratanakul2023guided}.

\subsection{Objective metrics}

\begin{figure}[t!]
\centering
  \includegraphics[width=\linewidth]{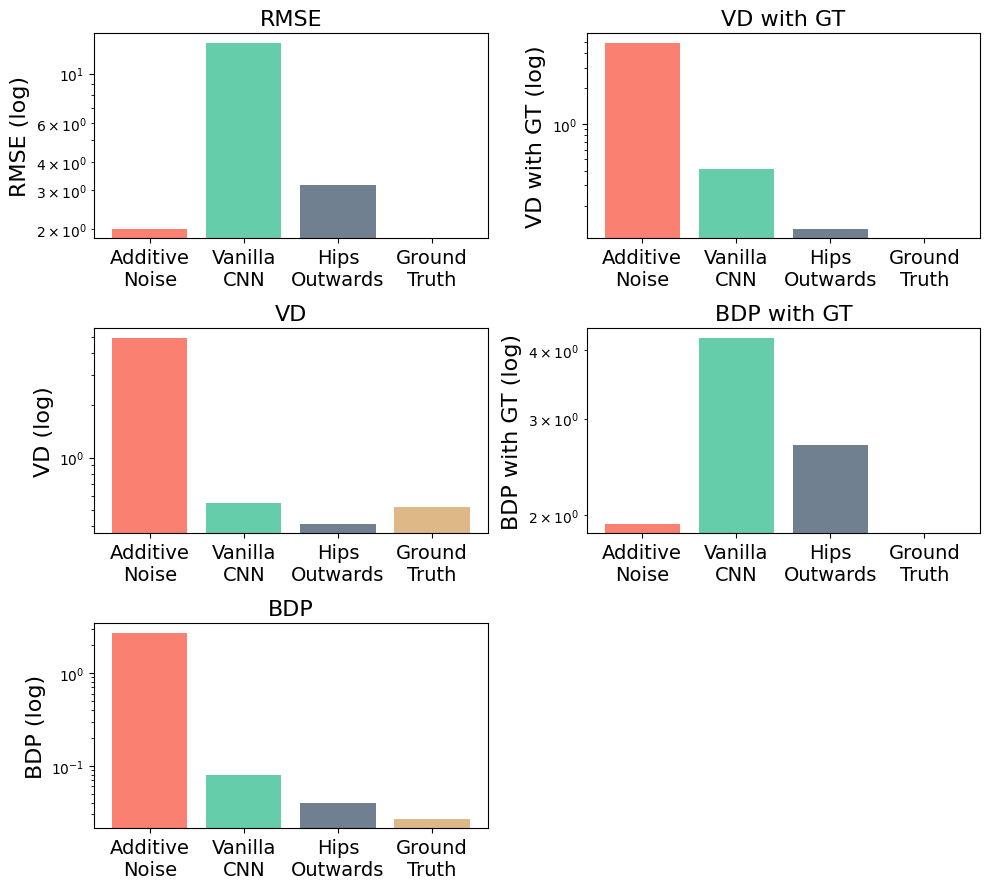}
  \caption{Results of the objective evaluation in cm with log-scale on the x-axis. RMSE stands for Root Mean Square Error, VD stands for Velocity Distance, GT stands for Ground Truth, and BDP stands for Bone Distance Preservation.  Lower values indicate better performance. System are sorted according to their ratings in the user study.}
  \label{fig:results_objective}
\end{figure}

The objective evaluation results in terms of average metric values are shown in Figure \ref{fig:results_objective}. We chose to visualize the metrics with a logarithmic scale as the Additive Noise condition creates outliers that impede readability.
The actual values are listed in Table~\ref{tab:objective} for comparison. We can see that RMSE is low in the Additive Noise condition even though the condition received the lowest ratings in the subjective evaluation. VD with GT sorts the conditions in the same order as the user study, while VD assigns a lower score to the Hips Outwards system than to the Ground Truth. As VD with GT, BDP also follows the same trend as the user study. BDP with GT on the other hand assigns lower scores (and thus better) to the Additive Noise condition than to the two ML systems.

Based on these observations, VD with GT and BDP without GT seem to be the most promising metrics as they rank the conditions in the same way as humans do. We will quantify this hypothesis in the next section.

\section{Correlation between objective metrics and subjective perception}

In order to understand the alignment between metrics and subjective quality perception,  we analyze the correlations on a stimuli level. Each sequence from each condition constitutes one stimulus. In our user study, most of the stimuli received at least five responses. We average the user ratings for each stimulus and correlate the result with the objective metrics calculated for the corresponding stimuli.

\begin{figure}[t]
\centering
  \includegraphics[width=1.05\linewidth]{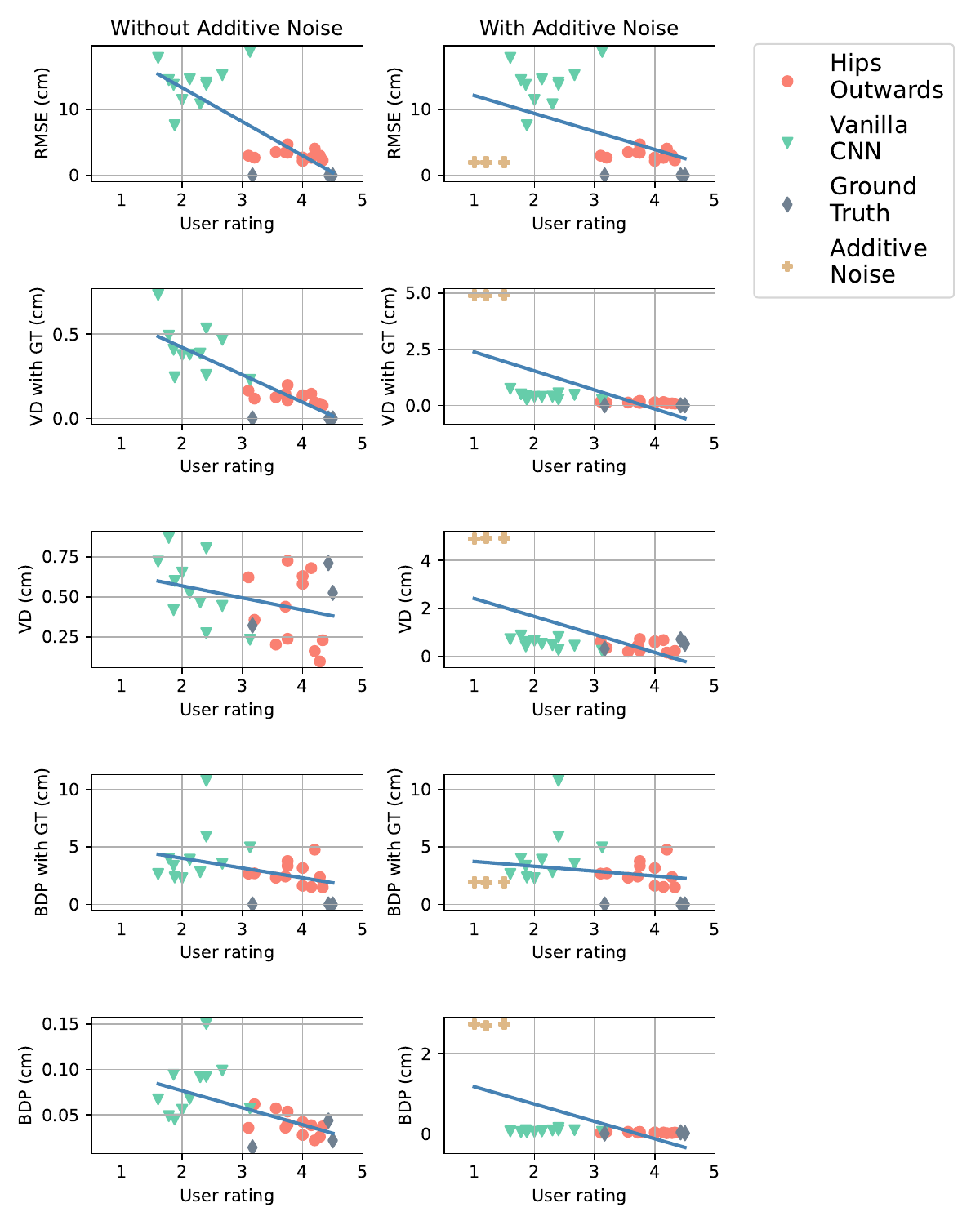}
  \caption{Metric values plotted against the average subjective perception score for each stimulus considered. The metrics include RMSE; VD with GT, VD, BDP with GT and BDP. For readibility, we plot the data without the Additive Noise condition in the left column and with the Additive Noise condition in the right column. }
  \label{fig:correlation}
\end{figure}

In Figure \ref{fig:correlation} we plot the metric values for each stimuli against the average user rating. As the Additive Noise condition sometimes has extreme values (e.g. in the case of VD with GT), we show the stimuli excluding Additive Noise in the left column and all stimuli in the right column. We also plot the trend lines for the respective datasets. One important observation is that the Additive Noise stimuli seem to be outliers in the case of RMSE as they do not follow the general negative trend. This is in line with our hypothesis that RMSE is no strong metric when it comes to marker reconstruction. Small random fluctuations have a low RMSE error but can result in poor perceputal quality. Similarly, BDP with GT seems to be a weak metric as the Additive Noise condition results in low values. 

 We assessed the metrics' quality by calculating Kendall’s $\tau$ correlations \cite{kendall1938new} between their rankings and that of the average user study ratings. This is similar to \cite{kucherenko2023evaluating}, but instead of ordering systems (of which we have only four), we order stimuli. For the metrics-based ordering, we sort the stimuli from smallest to largest values, as smaller values indicate better sequences. Conversely, for the user ratings, we sort the stimuli from highest to lowest, since higher values  correspond to better sequences. 

\begin{table}[!b]
    \centering
    \begin{tabular}{|l|c|c|c|}
    \toprule
        \textbf{Metric} & \textbf{Kendall} $\tau$ & \textbf{p-value} & \textbf{95\% CI $\tau$} \\ \midrule
RMSE &    0.2928 & 0.0267 & (0.065, 0.334) \\ 
VD with GT&  \textbf{0.7345} & 3e-08 & (0.618, 0.881)  \\ 
VD &     \textbf{0.3684} & 0.0052 & (0.149, 0.578)\\ 
BDP with GT & 0.1538 & 0.2443 & (-0.265, 0.283)\\ 
BDP &    \textbf{0.5464}&  3.358e-05& (0.44, 0.697)  \\ \bottomrule
    \end{tabular}
    \vspace{4mm}
    \caption{Kendall’s $\tau$ rank correlation coefficients and p-values for different objective metrics wrt the user study results. Bold font indicates statistically significant results (p<0.01). We also include 95\% confidence intervals for Kendall $\tau$   using non-parametric bootstrapping with 50,000 resamples, using the 2.5th and 97.5th percentile values. }
    \label{tab:correlation_metrics}
\vspace{1mm}
\end{table}
The results are presented in Table \ref{tab:correlation_metrics}.  We observe that VD with GT has a relatively high correlation, followed by BDP. Due to the small sample size, especially in the case of the Additive Noise and Ground Truth condition, our threshold for significance is 0.01. Thus we conclude that VD with GT, VD and BDP correlate significantly with subjective quality perception, while RMSE and BDP with GT do not. 

We also assess the inter-rater reliability using Krippendorff’s $\alpha$ \cite{krippendorff2018content} with an ordinal distance function. This statistic was chosen because our rating scale consists of ordered categories (e.g., 1 = low, 5 = high), and $\alpha$ accounts for the magnitude of disagreement between raters—not just its occurrence. Importantly, Krippendorff’s $\alpha$ can accommodate unequal numbers of ratings per item, which applied in our case: not all stimuli were rated by the same number of participants. Confidence intervals were obtained via non-parametric bootstrapping (50,000 resamples), resampling at the stimulus level, and calculated using the 2.5th and 97.5th percentiles of the resulting $\alpha$ distribution.  The resulting reliability estimate was $\alpha$ = 0.503 (95\% CI: 0.346, 0.624). This suggests a moderate level of inter-rater reliability across study participants.

\section{Discussion}

We compared objective metrics with subjective ratings by plotting them jointly and calculating correlations. The commonly used RMSE metric showed a notably weak correlation with subjective ratings (Kendall tau of 0.29). While Velocity Distance with Ground Truth achieved the highest Kendall tau of 0.73. 

We have learned a number of things from our experiments:

\begin{itemize}
    \item The commonly used objective metric, RMSE, should not be used as it does not correlate with subjective perception neither on the system level nor on the stimuli level.
    \item The rarely used (and adapted) objective metrics that take spatial correlation into account, Bone Distance Preservation without Ground Truth, correlates well with subjective perception both on the system level and on the stimuli level.
     \item The newly proposed objective metrics that take temporal correlation into account, Velocity Distance to the Ground Truth, correlates well with subjective perception both on the system level and on the stimuli level.
    \item Many of the manually cleaned-up sequences were rated to require further editing. Some of the lower rated sequences would require additional manual cleaning and can therefore not be considered as a clean ground truth. Depending on the quality of the manual cleaning, metrics that depend on ground truth sequences might therefore not always be reliable.
\end{itemize}

We also note that the evaluated objective metrics are differentiable and can be used as a loss function. Therefore, we suggest complementing MSE with other metrics both during training and evaluation.


Our results highlight the insufficiency of current metrics and the need of further research to develop reliable objective metrics that better align with subjective perception of quality in MoCap cleaning tasks. We also encourage the development of a task specific dataset with paired raw and clean MoCap data published with a clear evaluation protocol that divides the data according to difficulty levels in terms of temporal and spatial gaps. 

\section{Limitations}

Our experiments have several important limitations that we would like to highlight.

First, the study is limited by its scale and scope. We considered only one dataset, two systems, three sequences, and four segments from these sequences. As a result, the findings might not generalize to different datasets or models.

Additionally, our user study included only 15 participants, which may not fully eliminate the subjective aspect of the task. A larger participant pool would likely yield more robust and possibly slightly different results.   

Finally, the dataset used in this work is proprietary, and we cannot publicly release it, which limits reproducibility.

Despite these limitations, we believe this work remains valuable. It represents the first user study in this field, to our knowledge, and offers a critical evaluation of objective metrics.  
\section{Acknowledgements}

We are grateful to Tomas Tjarnberg and Alvaleia Thorsson for their data collection efforts on this project; to Harald Zlattinger and Chris Battson for their invaluable support, guidance, and resources; to Timur Solovev for his engineering support; to Jim Royal for help with crafting visuals; to Mónica Villanueva Aylagas, Héctor Leon, and Alessandro Sestini for feedback on the paper; and to Clement Lescalet, Jeremy Mathiesen, Aaron Punga, John Aguilar, Jae Lee, Kristy Clarke, Sam Kwok, Audrey Ito, Skyla Li, Vince Ng, Marcelo Sakai, Brennan Robinson, Vincent Hung, and Lucy Guo for participating in the user study.  

\balance

\bibliographystyle{ACM-Reference-Format}
\bibliography{sample-base}



\end{document}